 \documentclass[pmlr,wcp,twocolumn]{jmlr} 

\usepackage{booktabs}
\usepackage[load-configurations=version-1]{siunitx} 

\theorembodyfont{\upshape}
\theoremheaderfont{\scshape}
\theorempostheader{:}
\theoremsep{\newline}

\jmlrvolume{ML4H Extended Abstract Arxiv Index}
\jmlryear{2020}
\jmlrsubmitted{2020}
\jmlrpublished{}
\jmlrworkshop{Machine Learning for Health (ML4H) 2020} 

\usepackage{natbib}

\usepackage[utf8]{inputenc} 
\usepackage[T1]{fontenc}    
\usepackage{hyperref}       
\usepackage{url}            
\usepackage{booktabs}       
\usepackage{amsfonts}       
\usepackage{nicefrac}       
\usepackage{microtype}      
\usepackage{subcaption}

\usepackage{graphicx}

\title{\vspace{-0.4cm}Phenotyping Clusters of Patient Trajectories suffering from Chronic Complex Disease}

\author{%
\Name{Henrique Aguiar} \Email{henrique.aguiar@stcatz.ox.ac.uk}\\
\addr University of Oxford \\
\Name{Mauro Santos} \Email{mauro.santos@eng.ox.ac.uk}\\
\addr University of Oxford \\
\Name{Peter Watkinson} \Email{peter.watkinson@ndcn.ox.ac.uk} \\
\addr University of Oxford \\
\Name{Tingting Zhu} \Email{tingting.zhu@eng.ox.ac.uk} \\
\addr University of Oxford
\vspace{-1cm}}

\begin{document}

\maketitle

\begin{abstract}
  Recent years have seen an increased focus into the tasks of predicting hospital inpatient risk of deterioration and trajectory evolution due to the availability of electronic patient data. A common approach to these problems involves clustering patients' time-series information (such as vital sign observations) to determine dissimilar subgroups of the patient population. Most clustering methods assume time-invariance of vital-signs and are unable to provide interpretability in clusters that is clinically relevant, for instance, event or outcome information. In this work, we evaluate three different clustering models on a large hospital dataset of vital-sign observations from patients suffering from Chronic Obstructive Pulmonary Disease. We further propose novel modifications to deal with unevenly sampled time-series data and unbalanced class distribution to improve phenotype separation. Lastly, we discuss further avenues of investigation for models to learn patient subgroups with distinct behaviour and phenotypes.
\end{abstract}

\vspace{-0.6cm}
\section{Introduction}
\vspace{-0.2cm}

Diseases such as Chronic Obstructive Pulmonary Disease (COPD) and Cardiovascular Disease describe a broad spectrum of medical conditions, affecting a significant percentage of the overall population, \cite{adeloye2015global}. Such diseases are characterized by the existence of multiple distinct patient subgroups (corresponding to disease 'phenotypes'), largely distinguished by differences in pathology, as well as wildly dissimilar response to different treatments and medical interventions \cite{turner2015clinically}.

Vital-sign observation and Electronic Health Records (EHR) data  can be used to determine COPD phenotypes, \cite{pikoula2019identifying}, and ultimately decrease risk of deterioration of patients. Most common approaches in time-series analysis implement homogenization frameworks (\cite{zhang2019data} and \cite{rusanov2016unsupervised}), or modify the similarity measure between time-series \cite{giannoula2018identifying}. There are challenges applying these techniques to multi-dimensional, multi-modal and unevenly sampled vital-sign observation data. Furthermore, there is still a large gap in interpreting and understanding phenotypes in a clinically relevant fashion.

In this paper, we compare three distinct approaches to the problem of determining interpretable well-separated phenotypes on a large dataset of COPD patients: a 'classical' baseline based on the KMeans algorithm, an unsupervised deep learning model for topological interpretation of clusters, and a supervised-clustering approach through neural networks maximising cluster separability via clinical events. We also propose changes to loss functions within the latter and we show this innovation allows for better learning and phenotype separation.

\vspace{-0.4cm}
\section{Methods}
\vspace{-0.2cm}

In this section, we describe the clustering models we considered. 

\vspace{-0.2cm}
\paragraph{TimeSeriesKMeans (TSKM)} Introduced in \cite{JMLR:v21:20-091}, TSKM builds on the classical K-means algorithm by incorporating different similarity measures between time-series, including the "Euclidean" and "Dynamic-Time Warping" (DTW) (\cite{berndt1994using}) measures. The former corresponds to the standard baseline comparison (K-means algorithm on temporal data), while the latter has been shown to improve analysis of time-series. For each vital sign, TSKM iteratively computes cluster centroids by minimising intra-cluster distance and maximising distance between clusters. Cluster assignments for new patients follow a nearest-centroid approach. We observed experiments with DTW performing better than those with Euclidean similarity at capturing relevant clusters. We, therefore, considered DTW-TSKM as our time-series clustering baseline model.

\begin{figure}[b!]
		\vspace{-0.3cm}
		\centering
		\includegraphics[width=\linewidth, height=1.7cm]{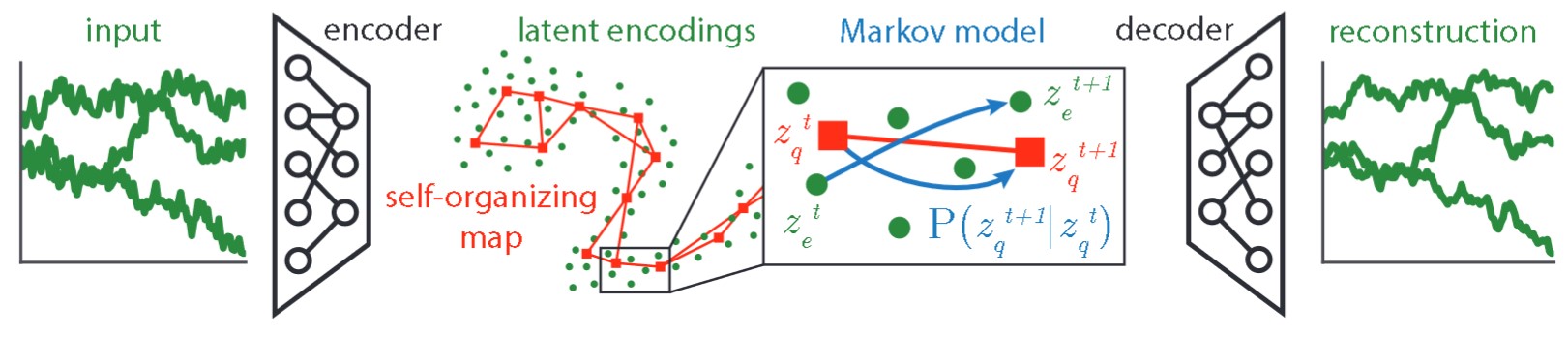}
		\caption{Diagram of SOM-VAE Model.}
		\label{fig:SOM_VAE_diagram}
\end{figure}

\vspace{-0.2cm}
\paragraph{SOM-VAE (\cite{fortuin2018som})} SOM-VAE  is a deep learning unsupervised model designed to learn topologically-interpretable discrete representations of time-series data. \figureref{fig:SOM_VAE_diagram} shows a diagram representation of the model. By considering probabilistic models and Variational Auto-Encoders (VAE), SOM-VAE computes latent representations of each time observation and assigns them to nodes on a Self-Organized Map (SOM) structure. On top of the SOM a Markov Model is added to learn patient trajectory evolution. Nodes in the SOM represent clusters, and multiple loss functions are proposed by the authors in order to provide relevant representations of time-series. We consider SOM-VAE in our experiments as a representative state of the art, purely unsupervised model for clustering multidimensional time-series data. While the model is designed to capture the optimal number of clusters, our experiments with SOM-VAE revealed high sensitivity to hyper-parameters, and difficulty learning relevant clusters.   

\begin{figure}[b!]
	\centering
	\includegraphics[width=\linewidth, height=2.7cm]{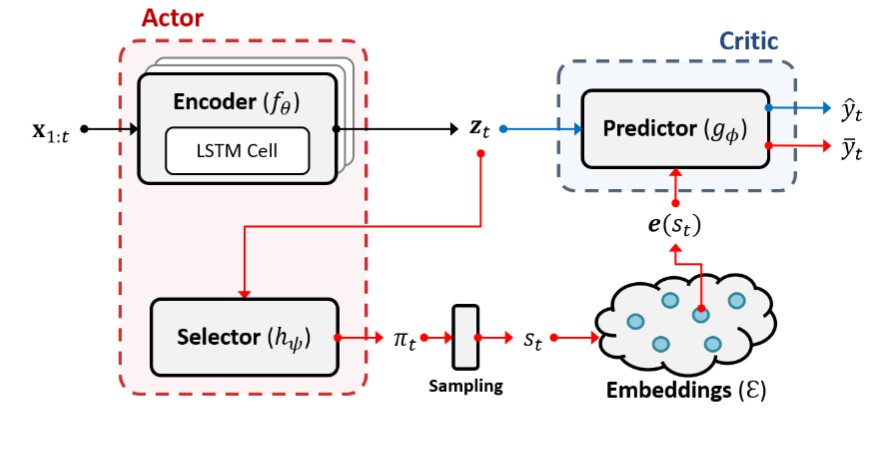}
	\caption{Diagram of AC-TPC Model.}
	\label{fig:AC_TPC_diagram}
\end{figure} 

\vspace{-0.2cm}
\paragraph{AC-TPC (\cite{lee2020temporal}) and new proposed loss} A sketch diagram for this probabilistic supervised clustering model is shown in \figureref{fig:AC_TPC_diagram}. AC-TPC utilises label information as part of the optimisation of the encoder and decoder, and then uses an actor-critic approach, \cite{konda2000actor} to learn relevant cluster embeddings and cluster representations. AC-TPC considers subsequences of data samples as part of its training process. We noticed difficulties learning relevant patient subgroups which we realized was due to the high class unbalance in our dataset, an issue common within medical datasets. We therefore propose a new weighted loss function to aid in learning. Following the original notation, let $\mathbf{y}_{t} = \left\{ y_{t}^{1}, ..., y_{t}^{C} \right\}$ represent the label of a patient sample in one-hot encoding format (we are assuming labels are assigned one out of $C$ possible classes) and let $\bar{\mathbf{y}}_{t} = \left\{\bar{y}_{t}^{1}, ..., \bar{y}_{t}^{C} \right\}$ be the predicted probability vector from the last softmax layer in the actor-critic pathway. We define:

\vspace{-0.4cm}
\begin{equation*}
    l(\mathbf{y}_{t},\bar{\mathbf{y}}_{t} ) = - \sum_{c=1}^{C} \frac{1}{\alpha_{c}} \mathbf{y}_{t}^{c} \log \bar{\mathbf{y}}_{t}^{c}
\end{equation*}
\vspace{-0.3cm}

where $\alpha_{c}$ is the proportion of patients with label $c$. Implementing this loss in both pre-training (where $\hat{\mathbf{y}}_{t}$, the predicted probability vector from the auto-encoder pathway is used instead of $\mathbf{y}_{t}$) and main training phases, we observed a much higher robustness of the model in not collapsing clusters and identifying multiple outcomes. In order to compare and interpret model clusters on testing data, we considered the cluster of an unseen patient trajectory to be the most common cluster for the patient over the last 48 hours. This value, based on clinical input, was selected as a representation of the patient's physiological state. We refer to AC-TPC$_w$, or simply AC-TPC, as the model implementing our proposed loss, while we denote the original as "AC-TPC old".
\vspace{-0.7cm}
\section{Dataset} 
\vspace{-0.2cm}

 Data was retrieved from a retrospective large database of routinely collected observations from concluded hospital admissions between 21st March 2014 and 31st March 2018 within the Hospital Alerting Via Electronic Noticeboard (HAVEN) project (REC reference: 16/SC/0264 and Confidential Advisory Group reference 08/02/1394). The database included the vital-sign measurements of adult patients admitted to four Oxford University Hospitals: the John Radcliffe Hospital, Horton General Hospital, Churchill Hospital, and the Nuffield Orthopaedic Hospital, collected by the System for Electronic Notification and Documentation (SEND, Sensyne Health), \cite{wong2015send}, \cite{pimentel2019comparison}.
 
 HAVEN includes 168,794 COPD patients with a total of 4,354,735 observations. In our analysis, outcomes consisted of either i) no event during hospital stay, leading to successful discharge from the hospital, or the first instance of one of three possible events: ii) unplanned entry to ICU, iii) cardiac arrest and iv) death. We used the protocol  defined in \cite{pimentel2019comparison} to identify COPD patients and respective outcomes. We consider $8$ vital signs: Heart Rate (HR), Respiratory Rate (RR), Diastolic Blood Pressure (DBP), Systolic Blood Pressure (SBP), Oxygen Saturation (SPO$_{2}$), Temperature (TEMP), 'Alert, Verbal, Pain, Unresponsive' (AVPU) and an estimation for Fraction of Inspired Oxygen (Estimated FIO$_{2}$, simply labelled FIO$_{2}$). Each observation consisted of numerical values for the corresponding vital-sign.

Pre-processing was based on work done in \cite{pimentel2019comparison}, typical outlier removal measures and clinical input. A full description of our pipeline is presented in \figureref{fig:pre-processing}, displayed in the Appendix. After processing, 16,469 patient last admissions remained. We note that the processed vital-signs had the same time index with respect to time to outcome. As input we considered solely observations up to 3 days before an outcome, while we visualized 7 days of data. We also remark that our dataset is very unbalanced - $93.90 \%$ of patients had a successful discharge.

\vspace{-0.5cm}
\section{Results}
\vspace{-0.2cm}
\begin{table}[b!]
	\centering
	\begin{tabular}{||p{0.28\linewidth} | p{0.17\linewidth} | 
			p{0.17\linewidth} | p{0.12\linewidth}||}
	\hline
	Model & AUROC & AUPRC & NMI \\
	\hline \hline
	TSKM & 0.439 & 0.062 & 0.077 \\
	SOM-VAE & 0.653 & 0.131 & 0.035 \\
	AC-TPC old & 0.776 & 0.338 & 0.155 \\
	AC-TPC$_w$ & \textbf{0.869} & \textbf{0.387} & \textbf{0.183} \\
	\hline \hline
	\end{tabular}
	\caption{Supervised Metric Scores}
	\label{table:supervised_Scores}
\end{table}

Model performance was evaluated using the following supervised scores: Area-under-the-Operating-Curve (AUROC), Area-under-Precision-Recall-Curve (AUPRC), Normalised Mutual Information (NMI). These scores help to determine how the patient groups learned by each model captured the different outcomes.

\begin{figure}[t!]
	\vspace{-0.3cm}
    \centering
    \includegraphics[width=\linewidth, height =0.37\textheight]{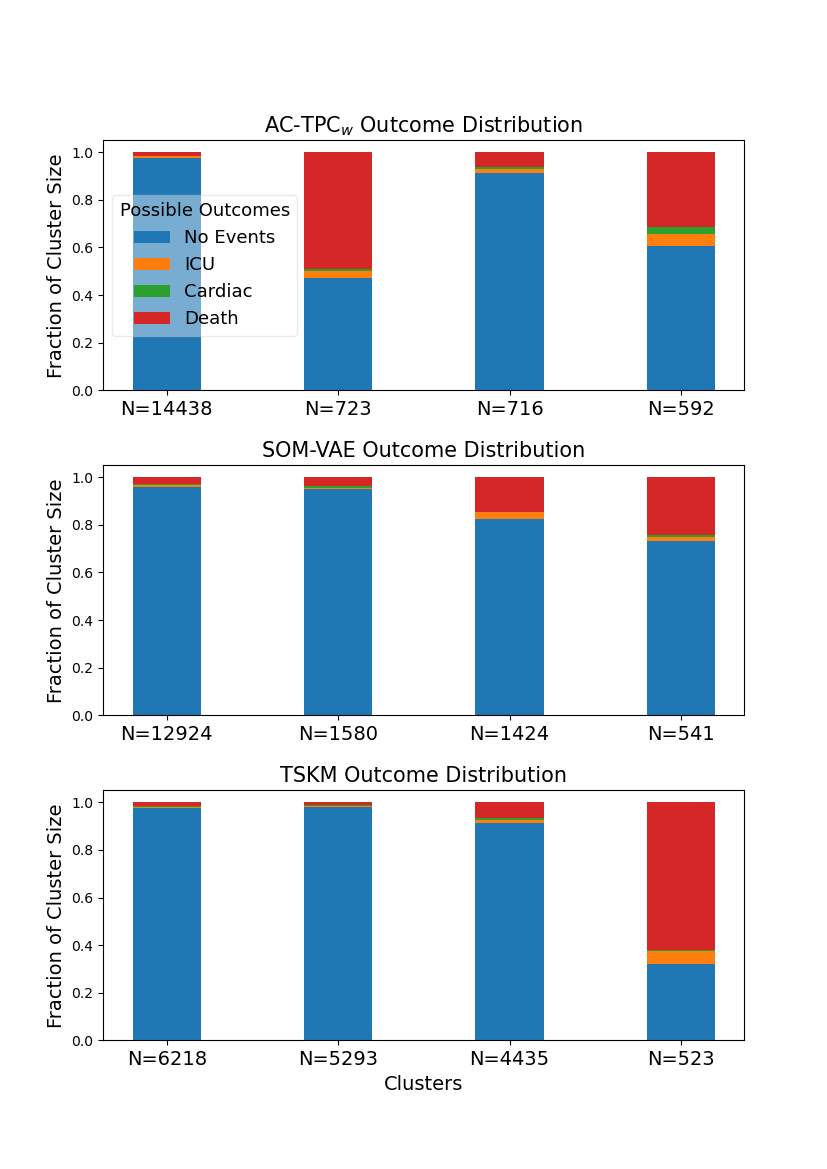}
    \vspace{-1.0cm}
    \caption{Outcome distribution of patients within each cluster (group) and model. Clusters are ordered from largest to smallest, and we note Clusters represent different population groups across models}
    
    \label{fig:outcome_piechart}
\vspace{-0.5cm}
\end{figure}

\begin{figure}[t!]
	\vspace{-0.2cm}
	\centering
	\includegraphics[height=0.37\textheight, width=\linewidth]{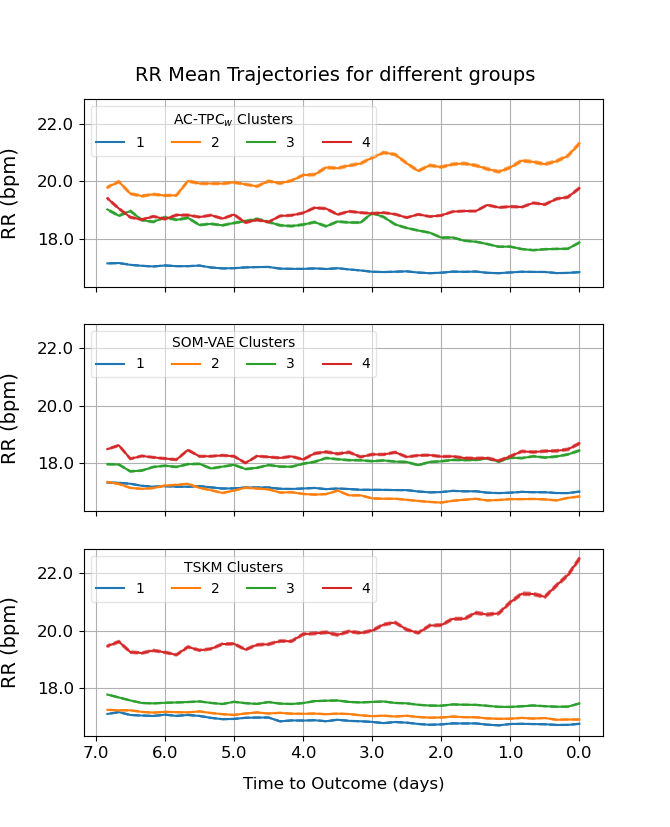}
	\vspace{-0.6cm}
	\caption{Respiratory Rate Mean Trajectory for each cluster and for each model. Clusters are ordered from largest to smallest, and number of patients displayed for each cluster.}
	\label{fig:RR_mean}
\vspace{-0.4cm}
\end{figure}

To determine the relevant number of clusters, $K$, for TSKM, we used an elbow method approach, \cite{thorndike1953belongs}, which led to selecting $K=4$ as the appropriate number of clusters. Both SOM-VAE and AC-TPC self-implemented methods for learning $K$ using initialisations procedures based on variants of the k-means++ algorithm. Both methods returned $K=4$ as the relevant number of clusters. Numeric results for the best experiments across all models (including a comparison with the original AC-TPC) are presented in Table \ref{table:supervised_Scores}. Furthermore, we plot outcome distribution of the patient population and number of patients in each cluster and model in \figureref{fig:outcome_piechart}. We also display the mean RR trajectory of each cluster for all models as a summary representation of each cluster's behaviour in \figureref{fig:RR_mean}.

Supervised score results indicate AC-TPC performs significantly better than other models. In \figureref{fig:RR_mean} we can observe mean trajectories for each AC-TPC cluster are more well-separated and variable over time, a desirable property in a clinical setting as vital-sign changes can be predicted in advance. Analyzing \figureref{fig:outcome_piechart} further shows AC-TPC obtains the highest dissimilarity in phenotype (with regard to outcome) across patient groups, as intended (e.g. despite the relatively small size, clusters 2 and 4 contain a proportionally large number of events, and around 40\% of all cardiac arrest outcomes can be found in Group 4).

\vspace{-0.6cm}
\section{Conclusion}
\vspace{-0.3cm}
In this work, we analyzed different clustering methodologies in a hospital dataset of multidimensional, multi-modal vital-sign observations, and evaluated their performance in obtaining well-separated clusters with distinct phenotypic characteristics. As part of our methodology, we introduced a novel loss function and obtained promising results with regards to the clustering of inhospital COPD patients. Looking forward, we aim to expand on previous work by incorporating other patient information, and identifying the individual contribution of vital signs and individual time-stamps to the cluster assignment, for instance, through attention mechanisms.

\bibliography{References.bib}

\section{Appendix}

\begin{figure}[h!]
	
	\centering
	\includegraphics[keepaspectratio, width = \linewidth]{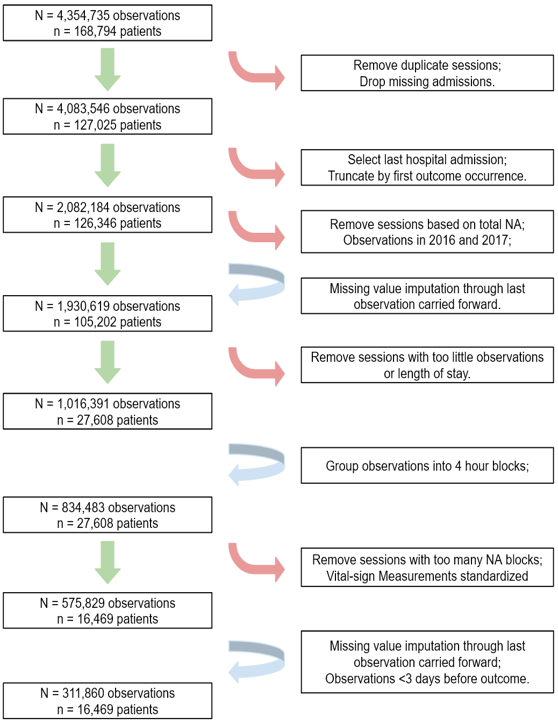}
	
	\caption{Sketch diagram of pre-processing steps on HAVEN Dataset}
	\label{fig:pre-processing}
	
\end{figure}

\end{document}